\documentclass[11pt]{article}

\usepackage[preprint]{acl}

\usepackage{times}
\usepackage{latexsym}

\usepackage[T1]{fontenc}

\usepackage[utf8]{inputenc}

\usepackage{microtype}

\usepackage{inconsolata}

\usepackage{tikz}
\usetikzlibrary{positioning} 
\usepackage{amsmath, amssymb} 
\usepackage{hyperref}         
\usepackage{booktabs}         
\usepackage{multirow}         
\usepackage{titlesec}
\usepackage{verbatim}
\usepackage{enumitem}
\usepackage{tabularx}
\usepackage{float}

\usepackage{graphicx}

\title{VotIE: Information Extraction from Meeting Minutes}

 \author{José Pedro Evans \\ INESC TEC and University of Porto \\  \textit{jose.joao@inesctec.pt} \And
 Luís Filipe Cunha \\ INESC TEC and University of Porto \\\textit{luis.f.cunha@inesctec.pt} \AND
 Purificação Silvano \\ INESC TEC and University of Porto \\ \textit{purificao.silvano@inesctec.pt} \And
 Alípio Jorge \\ INESC TEC and University of Porto\\ \textit{alipio.jorge@inesctec.pt} \AND
Nuno Guimarães \\ INESC TEC and University of Porto \\ \textit{nuno.r.guimaraes@inesctec.pt} \And
Sérgio Nunes \\ INESC TEC and University of Porto \\ \textit{sergio.nunes@inesctec.pt}  \AND
Ricardo Campos \\ INESC TEC and University of Beira Interior \\ \textit{ricardo.campos@inesctec.pt}}

\begin{document}
\maketitle

\vspace{-5.5cm}

\begin{abstract}

Municipal meeting minutes record key decisions in local democratic processes. Unlike parliamentary proceedings, which typically adhere to standardized formats, they encode voting outcomes in highly heterogeneous, free-form narrative text that varies widely across municipalities, posing significant challenges for automated extraction. In this paper, we introduce VotIE (Voting Information Extraction), a new information extraction task aimed at identifying structured voting events in narrative deliberative records, and establish the first benchmark for this task using Portuguese municipal minutes, building on the recently introduced CitiLink corpus.
Our experiments yield two key findings. First, under standard in-domain evaluation, fine-tuned encoders, specifically XLM-R-CRF,  achieve the strongest performance, reaching 93.2\% macro F1, outperforming generative approaches. Second, in a cross-municipality setting that evaluates transfer to unseen administrative contexts, these models suffer substantial performance degradation, whereas few-shot LLMs demonstrate greater robustness, with significantly smaller declines in performance. Despite this generalization advantage, the high computational cost of generative models currently constrains their practicality. As a result, lightweight fine-tuned encoders remain a more practical option for large-scale, real-world deployment. To support reproducible research in administrative NLP, we publicly release our benchmark, trained models, and evaluation framework.

\end{abstract}

\section{Introduction}

Natural Language Processing (NLP) has substantially advanced information extraction in governmental and administrative texts. However, most existing work focuses on document types with clear and standardized structures, such as legal documents \cite{LeNER-BR, ulyssesner} or parliamentary records \cite{Erjavec2023-parlamentary-ner}, where entities and decisions are explicitly marked and consistently formatted. In contrast, municipal council minutes record local decision-making in a highly variable, narrative form, which differs markedly across municipalities. Voting outcomes and policy decisions are often embedded in lengthy procedural descriptions and deliberations, without explicit boundaries or standardized templates. This heterogeneity poses significant challenges for automatic processing and large-scale analysis of local voting behavior \cite{Citizen_Participation_and_Machine_Learning,council_data_project}.




To address these challenges, we introduce VotIE (Voting Information Extraction), a task focused on extracting structured voting events from narrative deliberative records. VotIE formulates extraction as an event-argument identification problem: given a text segment describing a voting procedure, the task is to identify spans corresponding to voting event arguments, including the subject under deliberation, participants, and their respective stances (in favor, against, abstention, or absent), and vote counts. Building on this task formulation, we focus on defining the task and characterizing its difficulty. To this end, we introduce a benchmark built on the CitiLink corpus \cite{CitiLinkMinutes2025}, and use it to systematically evaluate feature-based models, fine-tuned Transformer encoders, and generative Large Language Models (LLMs), allowing us to characterize task difficulty and generalization behavior. Crucially, our experimental design extends beyond standard in-domain evaluation by explicitly assessing cross-municipality generalization, measuring model robustness to variation in local administrative writing styles.



Our results demonstrate that fine-tuned encoders, most notably XLM-R-CRF, achieve state-of-the-art performance on seen municipalities reaching 93.2\% exact-match macro F1. However, this strong in-domain performance does not transfer reliably to unseen municipalities, revealing a substantial generalization gap. In contrast, while generative LLMs underperform in the standard benchmark setting, they exhibit superior robustness under cross-municipality transfer, with smaller performance degradation. 

The main contributions of this work are as follows:
\begin{itemize}
    \item We formalize VotIE as a task for extracting structured voting events from narrative deliberative records and establish the first benchmark with standardized evaluation protocols, including cross-municipality generalization tests that quantify robustness to administrative writing-style variation
    \item We provide a comprehensive set of baselines spanning both discriminative and generative extraction paradigms.
    \item We publicly release our benchmark splits, fine-tuned models on all six municipalities, training code, and evaluation scripts to support reproducible research and future work in administrative NLP. \footnote{\url{https://huggingface.co/Anonymous3445/XLM-RoBERTa-CRF-VotIE}}
\end{itemize}

\section{Related Work}

Previous research has explored information extraction from administrative texts, particularly legal documents and parliamentary proceedings, yet the structured extraction of voting events from municipal council minutes has received little attention. In legal corpora, such as UlyssesNER-Br \cite{ulyssesner} and LeNER-Br \cite{LeNER-BR}, methods have primarily targeted general entity types (e.g., persons, organizations, legislation) rather than complex event structures. Feature-based approaches have also been applied to Portuguese municipal minutes for specific entities, such as subsidies \cite{portuguese-municipal-minutes-2010}. However, these methods did not target voting events, and they relied on traditional, context-independent features rather than modern context-aware embeddings.

Research on parliamentary and meetings corpora has focused on tasks such as topic modeling \cite{Erjavec2023-parlamentary-ner}, summarization \cite{hu-etal-2023-meetingbank,2022-elitr}, or decision detection \cite{AMI_meeting_corpus}. In contrast to both parliamentary records, where votes are structured, and conversational meeting transcripts, where decisions are implicit, municipal council minutes often present formal voting events embedded within narrative prose. 
 
Existing methods, including sequence labeling approaches, are limited in their ability to extract explicit voting elements, whereas generative Large Language Models (LLMs) which have demonstrated promise for zero-shot Information Extraction \cite{wadhwa-etal-2023-revisiting}, have yet to be systematically applied to administrative records. 

\section{VotIE Task Definition}

\begin{figure}[ht!]
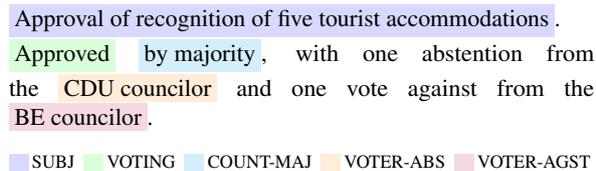
 
\small
\fboxsep=2pt

\noindent
\colorbox{blue!15}{Approval of recognition of five tourist accommodations}.
\colorbox{green!15}{Approved} \colorbox{cyan!15}{by majority}, with one 
abstention from the \colorbox{orange!15}{CDU councilor} and one vote against 
from the \colorbox{purple!15}{BE councilor}.

\medskip

{\scriptsize\centering\noindent
\colorbox{blue!15}{\rule{0pt}{1ex}\rule{1ex}{0pt}}~SUBJ\hspace{2pt}
\colorbox{green!15}{\rule{0pt}{1ex}\rule{1ex}{0pt}}~VOTING\hspace{2pt}
\colorbox{cyan!15}{\rule{0pt}{1ex}\rule{1ex}{0pt}}~COUNT-MAJ\hspace{2pt}
\colorbox{orange!15}{\rule{0pt}{1ex}\rule{1ex}{0pt}}~VOTER-ABS\hspace{2pt}
\colorbox{purple!15}{\rule{0pt}{1ex}\rule{1ex}{0pt}}~VOTER-AGST
}
\caption{The VotIE task: extracting structured voting event arguments from narrative municipal minutes.}
\label{fig:votie_task}
\end{figure}

We formalize the VotIE task as the extraction of structured voting events from narrative records of deliberative decision-making. Given a document represented as a sequence of tokens $\mathbf{X} = (w_1, \dots, w_n)$, the goal is to extract the set of all voting-related spans $\mathcal{S} = \{(s_j, e_j, t_j)\}$, where $s_j$ and $e_j$ are the start and end character offsets, and $t_j$ is the argument type. Unlike standard named entity recognition, which identifies mentions of categorical entities (e.g., persons, organizations, locations), VotIE focused on role-specific event arguments that together represent a structured voting event, as illustrated in Figure \ref{fig:votie_task}. A complete voting event includes at least one subject span and one voting span, optionally augmented with voter and count spans. While our extraction targets individual spans, their collective assembly enables reasoning over complete vote outcomes. This assembly, however, falls outside the scope of this paper and is left for future work.

\paragraph{Schema Definition:} Our extraction schema $\mathcal{T}$ defines four categories of voting event arguments. \textbf{Subject} captures the matter under deliberation, ranging from short noun phrases (e.g., "Budget approval") to complex multi-clause descriptions (e.g., "Approval of the protocol with the association for the implementation of social programs in the historical district"). \textbf{Voter Roles} identify participants according to their voting stance, with four subtypes: favor, against, abstention, and absent. \textbf{Voting Evidence} marks the decision action (e.g., "approved", "deliberated"), while \textbf{Counting Expressions} denote vote aggregation (e.g., "unanimously", "by majority").




\section{Experimental Setup}
\label{sec:experimental_setup}

\paragraph{Annotated Dataset:} We use the CitiLink-Minutes corpus \cite{CitiLinkMinutes2025}, which comprises 120 meeting minutes (2021–2024) from six Portuguese municipalities (M01--M06). The corpus contains 2,879 annotated segments (991,516 tokens), with a reported inter-annotator agreement (Cohen's $\kappa$) of 0.91 for the voting layer. All sensitive personal information, including names, addresses, and identifiers, was manually anonymized. Regarding voting related entities, the dataset contains 9,721 annotated entities. The distribution across argument types is as follows: Voting (27.24\%), Subject (23.13\%), Voter-Favor (16.97\%), Counting-Unanimity (16.15\%), Voter-Abstention (10.72\%), Counting-Majority (2.96\%), Voter-Against (1.82\%), and Voter-Absent (1.01\%). For the standard benchmark, we used a stratified 70/15/15 split to ensure balanced representation across both entity types and municipalities. For cross-municipality evaluation, we perform leave-one-municipality-out (LOMO) experiments, training models on five municipalities and testing on the held-out sixth. The annotation guidelines, the dataset and the training/evaluation code are publicly available on the GitHub repository. \footnote{\url{https://github.com/Anonymous3445/VotIE}}

\paragraph{Extraction Paradigms:} We investigate two distinct computational  approaches to extract $\mathcal{S}$ from input $X$. In the \textbf{Sequence Labeling} paradigm, we employ standard BIO encoding, mapping input tokens $X$ to a label sequence $Y$ via encoder representations $H$. The model maximizes $P(Y|X)$ using either a linear classifier or a Conditional Random Field to capture transition dependencies ~\cite{Lafferty2001}. Spans are reconstructed by aggregating contiguous tags. In the \textbf{Generative Extraction} paradigm, the task is formulated as text-to-structure generation, where the model estimates $P(\mathbf{Z} | \mathbf{X})$, and $\mathbf{Z}$ is a structured string (e.g., JSON) representing the textual content of the spans. Generated strings are then aligned back to the source $\mathbf{X}$ to recover the start and end indices $(s, e)$, using proximity-based disambiguation when multiple occurrences exist.


\paragraph{Evaluation Metrics.} We report macro-averaged entity-level precision, recall, and F1. Our primary metric uses exact match following CoNLL conventions \cite{tjong2003introduction}, which requires both identical boundaries and type agreement. We also compute partial match F1 following SemEval-2013 conventions \cite{segura-bedmar-etal-2013-semeval}, which requires type agreement and boundary overlap. For discriminative models, BIO predictions are converted to spans and evalutated using seqeval \cite{seqeval}, while for generative models, character offsets are recovered via exact substring matching \cite{wang2023gptner}.

\paragraph{Training Setup:} We evaluate seven architectures across two paradigms, as listed in Table~\ref{tab:model_details}. Training follows standard practices for Portuguese sequence labeling \cite{Bert-CRF-Souza}: AdamW optimization (learning rate \(5 \times 10^{-5}\)), class-weighted cross-entropy loss, maximum 10 epochs with early stopping (patience 3), and 512-tokens sliding windows with 50-token overlap. Invalid BIO transitions are corrected post-decoding.
 For generative extraction, we test Gemini 2.5 Pro \cite{gemini25} 
in both zero-shot and few-shot (5 examples) configurations, employing schema-constrained decoding via native JSON APIs. 
Greedy decoding ensures deterministic generation. Generated spans undergo exact substring alignment with proximity-based disambiguation for duplicate occurrences. All the used prompts are available on the project repository. 

\begin{table}[ht!]
\caption{Model architectures and implementation details.}
\label{tab:model_details}
\centering
\small
\setlength{\tabcolsep}{3pt} 
\begin{tabularx}{\columnwidth}{l >{\raggedright\arraybackslash}X}
\toprule
\textbf{Model} & \textbf{Architecture details} \\
\midrule
CRF & Linear-chain CRF + token/domain features \cite{Lafferty2001} \\
BiLSTM-CRF & FastText + CharCNN + CRF layer \cite{huang2015bidirectionallstmcrf} \\
BERTimbau-Large & Encoder + Linear/CRF head \cite{souza2020bertimbau} \\
mDeBERTa-v3 & Encoder + Linear/CRF head \cite{2021debertav3} \\
XLM-RoBERTa & Encoder + Linear/CRF head \cite{2019XLM-Roberta} \\
\midrule
Gemini 2.5 Pro & API + Schema-constrained decoding \cite{gemini25} \\
\bottomrule
\end{tabularx}
\end{table}

\section{Results and Discussion}
\label{sec:results}

\paragraph{Standard Benchmark Performance.}
Table~\ref{tab:main_results_compact} reports performance on the test set. The multilingual \textbf{XLM-R-CRF} achieves the highest Exact Match F1 with 93.2\%, surpassing the Portuguese-specific BERTimbau-CRF by 6.2 points. The effect of the CRF layer is architecture-dependent: it provides a 8.3-point F1 gain for XLM-R, whereas for DeBERTa it primarily balances precision and recall. A substantial 29.1 point gap separates the best encoder from the top-performing generative model, which decreases to 16.4 points under Relaxed Match criteria. This disparity suggests that while LLMs are effective at identifying relevant semantic regions, they struggle to match the strict token-level boundaries required for this extraction task.

\begin{table}[ht!]
\centering
\caption{Macro averaged results of all baselines on the standard benchmark.}
\label{tab:main_results_compact}
\footnotesize
\setlength{\tabcolsep}{0pt} 
\begin{tabular*}{\columnwidth}{@{\extracolsep{\fill}} l ccc c ccc @{}}
\toprule
& \multicolumn{3}{c}{\textbf{Exact Match}} && \multicolumn{3}{c}{\textbf{Relaxed Match}} \\
\cmidrule{2-4} \cmidrule{6-8}
\textbf{Model} & \textbf{P} & \textbf{R} & \textbf{F1} && \textbf{P} & \textbf{R} & \textbf{F1} \\
\midrule
DeBERTa-CRF      & 89.4 & 91.6 & 90.4 && 95.0 & 93.5 & 94.1 \\
DeBERTa-Lin.     & 86.7 & 94.9 & 90.4 && 92.0 & \textbf{98.3} & 94.9 \\
XLM-R-CRF        & \textbf{91.0} & \textbf{95.6} & \textbf{93.2} && \textbf{95.2} & 96.8 & \textbf{95.9} \\
XLM-R-Lin.       & 78.5 & 94.0 & 84.9 && 84.8 & 97.5 & 90.2 \\
BERTimbau-CRF    & 82.0 & 93.0 & 87.0 && 88.1 & 96.3 & 91.9 \\
BiLSTM-FT        & 74.3 & 68.3 & 70.2 && 83.4 & 69.4 & 74.3 \\
CRF              & 84.7 & 72.7 & 75.1 && 86.9 & 74.3 & 77.0 \\
\midrule
Gemini-2.5 (5s)  & 62.7 & 65.9 & 64.1 && 76.8 & 82.8 & 79.5 \\
Gemini-2.5 (0s)  & 55.6 & 50.0 & 52.3 && 81.9 & 73.1 & 76.8 \\
\bottomrule
\end{tabular*}
\end{table}


\noindent\textbf{Cross-Municipality Generalization.}
To assess robustness to lexical and structural variation across municipalities, we conduct leave-one-municipality-out (LOMO) experiments. For the generative approach, examples from the held-out municipality were excluded from the few-shot prompts. While XLM-R-CRF achieves the highest performance on the standard benchmark, LOMO experiments were conducted with DeBERTa-CRF due to computational constraints. Future work should evaluate XLM-R's cross-municipality generalization. Table~\ref{tab:cross_muni} reports macro F1 scores on held-out municipalities. These results highlight a clear trade-off between in-domain performance and generalization. The fine-tuned encoder struggles on unseen writing styles, particularly in M06, suggesting overfitting to specific phrasing patterns present in the training data. In contrast, Gemini exhibits superior cross-municipality robustness, consistently outperforming the encoder on held-out municipalities. This indicates that the LLM captures the underlying meaning of the voting events rather than relying on surface-level lexical cues, making it a more reliable choice for deployment in regions with no available training data.

\begin{table}[ht!]
\centering
\small
\setlength{\tabcolsep}{3pt} 
\caption{Cross-municipality generalization (LOMO). Macro F1 scores on held-out municipalities.}
\label{tab:cross_muni}
\begin{tabular}{llcccccc}
\toprule
\textbf{Metric} & \textbf{Model} & \textbf{M01} & \textbf{M02} & \textbf{M03} & \textbf{M04} & \textbf{M05} & \textbf{M06} \\
\midrule
\multirow{2}{*}{\textbf{Exact}} & DeBERTa & 46.5 & 65.2 & 62.4 & 61.3 & 64.1 & 17.5 \\
& Gemini  & \textbf{66.8} & \textbf{69.0} & \textbf{69.7} & \textbf{62.0} & \textbf{76.0} & \textbf{45.4} \\
\midrule
\multirow{2}{*}{\textbf{Relaxed}} & DeBERTa & 68.5 & 71.1 & 80.3 & 74.3 & 66.7 & 31.3 \\
& Gemini  & \textbf{80.5} & \textbf{77.9} & \textbf{80.7} & \textbf{94.8} & \textbf{85.8} & \textbf{63.6} \\
\bottomrule
\end{tabular}
\end{table}

\section{Error Analysis}
\label{sec:error_analysis}

Table~\ref{tab:error_classification} categorizes errors into Missing (MIS), Spurious (SPU), Boundary (BND), and Type (TYP). Among the models, XLM-R-CRF is the most robust, yielding the lowest total error count and a relatively balanced distribution across error types. DeBERTa-CRF exhibits a conservative bias, minimizing spurious predictions (14\%) but suffering from higher missing rates (53\%). In contrast, the generative model shows the opposite pattern: with high segment-level recall but substantial over-generation, resulting in 40\% of its errors being spurious. Boundary errors remain a consistent challenge across all architectures, primarily due to the length and syntactic complexity of Subject spans.

\begin{table}[ht!]
\centering
\footnotesize
\caption{Error classification analysis in the standard benchmark for discriminative and generative models under exact match criteria.}
\label{tab:error_classification}
\begin{tabular}{@{}lrrrrr@{}}
\toprule
\textbf{Model} & \textbf{Total} & \textbf{MIS} & \textbf{SPU} & \textbf{BND} & \textbf{TYP} \\
\midrule
XLM-R-CRF   & \textbf{124} & \textbf{47} & 23 & 46 & 8 \\
DeBERTa-CRF & 147 & 78 & \textbf{21} & \textbf{42} & \textbf{6} \\
Gemini      & 601 & 145 & 243 & 206 & 7 \\
\bottomrule
\end{tabular}
\end{table}

\section{Conclusion}
\label{sec:conclusion}
We introduced VotIE, a novel task and benchmark for extracting structured voting events from deliberative records. Our experiments across discriminative and generative paradigms yield two main insights. First, fine-tuned encoders remain superior for precise span extraction, with XLM-R-CRF achieving state-of-the-art (93.2\% F1), significantly outperforming few-shot LLMs, which struggle with strict boundary alignment. Second, while these encoders perform well in-domain, they degrade sharply on unseen municipalities due to overfitting to specific writing styles, whereas LLMs demonstrate superior stability, making them better suited for low-resource or zero-shot settings. Although generative models offer superior adaptability, their tendency toward spurious extraction, combined with high inference costs, makes lightweight discriminative encoders the more practical choice for large-scale applications where annotated data is available.

\section{Limitations}
\label{sec:limitations}

 \paragraph{Task Formulation Limitations:} While VotIE formulates voting extraction as a span-identification task, fully reconstructing complex administrative procedures requires moving beyond individual argument extraction. Our current approach treats all voting arguments within a segment as a single event structure. However, meeting minutes occasionally describe multiple distinct votes within the same segment. For example, a motion and an amendment voted on continuously. In such cases, a sequence labeling approach may successfully identify all entities but cannot explicitly link each VOTER span to its corresponding SUBJECT or OUTCOME span. Future work should explore Relation Extraction (RE) or Event Extraction paradigms to handle overlapping or interrelated events more accurately.
 
\paragraph{Coreference Resolution:} Our evaluation focuses on segment-level extraction, treating each segment in isolation, which ignores cross-segment context and, importantly, coreference. In administrative texts, voters are frequently referred to by role (e.g., "the Councilor") or pronominal references rather than proper names, often relying on earlier context or external records. Consequently, while models can detect mentions of voters, they do not resolve these mentions to specific individuals in a knowledge base. Integrating coreference resolution is therefore a necessary next step to support detailed analysis of individual voting behavior.

\paragraph{Baselines and Experiments:} Our generative AI experiments prioritized an efficient and easily-deployable model (Gemini 2.5 Pro) reflecting the resource constraints typical of local government setting. While larger state-of-the-art models (e.g., GPT-5, Claude 4.5, or Llama 4.0) might yield superior performance, their computational cost and latency currently limit scalability for processing decades of municipal recods across thousands of municipalities. A key observation of our experiments was is that the Leave-One-Municipality-Out (LOMO) experiments (Table \ref{tab:cross_muni}) reveal performance degradation when models encounter new municipalities. Although encoder models remain robust, variation in writing style and lexical choices introduce domain shifts that these models cannot fully overcome. Adrressing this may requires larger, geographically diverse datasets or unsupervised domain adaptation techniques to improve generalization.

\paragraph{Language:} Finally, this study focuses exclusively on Portuguese municipal records. While the VotIE schema (Subject, Voter, Stance, Count) is theoretically language-agnostic, the syntactic and stylistic patterns of administrative Portuguese differ significantly from other languages. Further research is needed to assess whether the architectural trends observed here, specifically the superior performance of fine-tuned encoders over LLMs, generalize to other linguistic contexts.

\section{Ethical Considerations}
\label{sec:ethical}

\paragraph{Intended Use and Misuse Potential:} The primary goal of VotIE is to enhance transparency in local governance by making unstructured data more accessible. However, applying automated information extraction to public administrative records carries a risk of unintended misinformation. As noted in our Error Analysis section \ref{sec:error_analysis}, neither fine-tuned encoders nor generative LLMs achieve perfect accuracy, with the latter showing a pronounced tendency towards spurious generation. In a real-world civic technology context, extraction errors that misattribute a vote or stance to a political actor could misrepresent their public position, potentially damaging their reputation and distorting public perception. Consequently, we emphasize that systems built on VotIE should be deployed as human-in-the-loop support tools for journalists and civil society organizations, rather than fully autonomous authorities on legislative truth. They purpose is to assist in navigating voluminous records, not to replace official, legally binding minutes.

\paragraph{Privacy and Data Usage:} The dataset used in this work comprises public municipal records, which are, by law, open to the public. However, to mitigate potential harm to private citizens occasionally mentioned in these proceedings (e.g., petitioners), we relied on the anonymized version of the CitiLink corpus, in which all personal identifiers were manually removed prior to model training.

\paragraph{Usage of AI:} In accordance with conference policy, we state that AI assistants were used to enhance clarity and readability, improving language flow and presentation. Moreover, code assistance systems were used to write some subroutines. All scientific claims, experimental designs, and data analyses are the original work of the authors.

\section*{Acknowledgments}

This work is funded by Component 5 - Capitalization and Business Innovation, integrated in the Resilience Dimension of the Recovery and Resilience Plan within the scope of the Recovery and Resilience Mechanism (MRR) of the European Union (EU), framed in the Next Generation EU, for the period 2021 - 2026, within project CitiLink, with reference 2024.07509.IACDC (\url{https://doi.org/10.54499/2024.07509.IACDC}).

\bibliography{custom}



\end{document}